\documentclass{article}

\usepackage[a4paper,margin=1in]{geometry}

\usepackage{graphicx}
\usepackage{amsmath, amssymb}
\usepackage{booktabs}
\usepackage{array}
\usepackage{url}
\usepackage{enumitem}
\usepackage[numbers,sort&compress]{natbib}
\usepackage[hidelinks]{hyperref}

\title{A Metamorphic Artificial Age Score (AAS) Decision-Support Prototype for Flight-Log-Based Drone Propeller Health Monitoring}

\author{
\large Seyma Yaman Kayadibi\\[0.4em]
\small Institute for Sustainable Industries and Liveable Cities (ISILC)\\
\small Victoria University, Melbourne, Australia\\
\small \texttt{seyma.yamankayadibi@live.vu.edu.au}
}

\date{}

\begin{document}

\maketitle

\begin{abstract}
Drone propeller faults can create safety and reliability risks when their effects are distributed across multiple flight-log channels rather than appearing as a single diagnostic signal. This paper proposes a Metamorphic Artificial Age Score (AAS) decision-support prototype for flight-log-based drone propeller health monitoring. Using selected historical real flight logs from the 2024 DronePropA public dataset, the framework computes six health-related indicators from raw MATLAB \texttt{.mat} matrices: trajectory tracking error, attitude instability, thrust-command burden, motor-command imbalance, ESC-command instability, and battery-level stress. These indicators are normalized relative to a healthy baseline and evaluated through candidate scoring policies, metamorphic adequacy relations, and a redundancy-adjusted AAS formulation. Here, AAS is used as a structural policy-adequacy and burden measure rather than as a chronological age measure. A controlled retrospective evaluation was performed using one healthy baseline and three defective propeller cases under the same speed profile and trajectory. The healthy case was assigned to routine monitoring. The Severity 1 case was dominated by ESC-command instability and assigned to maintenance review. The Severity 2 case reached maximum motor-command and ESC-command burden, while the Severity 3 case reached maximum trajectory tracking error; both triggered mandatory inspection. The results show that propeller fault effects may appear through different operational channels, supporting the need for a multi-indicator decision-support layer for post-flight maintenance prioritization and autonomous-system oversight.
\end{abstract}


\section{Introduction}

Unmanned aerial vehicles (UAVs) are increasingly used in operational domains where reliability, safety, and maintainability are essential, including logistics, inspection, agriculture, emergency response, and public-safety applications. In these settings, the condition of core flight components becomes a critical operational concern. Propellers are particularly important because they directly influence thrust generation, flight stability, trajectory tracking, and control responsiveness. Propeller faults such as cracks, edge damage, surface cuts, and imbalance may alter aerodynamic and mechanical behaviour even when the drone remains operational. The DronePropA data article emphasizes that propeller reliability problems can generate safety risks and financial losses, motivating the development of effective health-monitoring and predictive-maintenance methods \cite{ismail2025}.

A central challenge in drone propeller health monitoring is that fault effects may be distributed across several flight-log channels rather than appearing as a single diagnostic signal. The DronePropA public dataset, released in 2024 and later described in a \textit{Data in Brief} article, provides historical real flight-log data for healthy and defective propeller conditions \cite{ismail2024,ismail2025}. It contains 130 flight sequences across three fault types, three severity levels, two speed profiles, and five trajectory patterns \cite{ismail2025}. The associated MATLAB flight logs include measured and reference positions, orientation-related signals, thrust commands, motor commands, electronic speed controller (ESC) commands, and battery-level information. This structure makes DronePropA suitable not only for fault classification, but also for examining how flight-log-derived indicators can support interpretable maintenance decisions.

Existing UAV health-monitoring studies have commonly focused on fault detection, diagnosis, and classification. Vibration-based approaches are especially relevant because propeller and rotor faults can produce measurable changes in inertial or accelerometer signals. Baldini et al. proposed a real-time propeller fault-detection method for multirotor drones based on vibration-data analysis, demonstrating the value of signal-based monitoring for identifying propeller faults \cite{baldini2023}. Ghazali and Rahiman examined vibration-based drone fault detection using artificial-intelligence methods, further supporting the role of data-driven approaches in UAV health monitoring \cite{ghazali2022}. These studies show that fault-related signals can be extracted from drone data; however, classification alone does not fully address the downstream decision-support problem.

The decision-support problem differs from the classification problem. A classifier may label a drone as healthy or faulty, but maintenance teams also need to understand which operational indicator drives the recommendation, whether the evidence is strong enough to justify inspection, and whether the scoring policy behaves consistently under meaningful input transformations. For example, an increase in trajectory tracking error should not be ignored; motor-command imbalance should not be fully masked by a decrease in a less relevant indicator; and battery-level variation alone should not be interpreted as propeller criticality when the main propeller-dynamics indicators remain low. These are structural adequacy questions concerning how a decision-support policy responds to changes in the input space.

This study is guided by the following research question: can selected DronePropA flight-log signals be transformed into an interpretable post-flight decision-support output using a metamorphic Artificial Age Score framework? This question is important because drone propeller health monitoring is not only a fault-detection problem. In operational settings, maintenance teams need to know why a case requires routine monitoring, maintenance review, or mandatory inspection. A binary healthy/faulty classification may identify the presence of a fault, but it may not explain which flight-log channel drives the decision or whether the scoring mechanism behaves consistently under expected changes in the input indicators.

The significance of this study lies in its attempt to connect three layers that are usually treated separately: flight-log-derived feature extraction, decision-support scoring, and policy-adequacy evaluation. The proposed prototype uses real DronePropA flight-log data to compute six health-related indicators, then evaluates candidate scoring policies through metamorphic adequacy relations and a redundancy-adjusted AAS formulation. This makes the framework useful not as a certified diagnostic system, but as a transparent proof-of-concept decision-support layer for post-flight maintenance prioritization.

The study is also significant because it adapts AAS from a behavioural or structural aging measure into a policy-adequacy measure for drone health monitoring. In this setting, AAS does not measure chronological age. Instead, it measures the structural inconsistency of a scoring policy when the input vector is transformed according to health-monitoring expectations. This allows the proposed framework to evaluate not only the final propeller-health score, but also the reliability of the policy that produces that score.

This paper proposes a Metamorphic Artificial Age Score (AAS) decision-support prototype for flight-log-based drone propeller health monitoring. The proposed framework transforms selected historical real DronePropA flight logs into six normalized health-related indicators: trajectory tracking error, attitude instability, thrust-command burden, motor-command imbalance, ESC-command instability, and battery-level stress. These indicators are computed from raw MATLAB \texttt{.mat} time-series matrices rather than manually assigned. They are then evaluated through candidate scoring policies, metamorphic adequacy relations, and a redundancy-adjusted AAS formulation. The aim is to provide an interpretable post-flight decision-support layer for propeller health monitoring, maintenance prioritization, and autonomous-system oversight workflows.

In this study, Artificial Age Score is used in a structural rather than chronological sense. In the original AAS formulation, artificial age does not refer to elapsed time, biological aging, or calendar age. Instead, it represents behavioural or structural burden through consistency loss, redundancy adjustment, and logarithmic penalty \cite{kayadibi2026}. Here, this logic is transferred from AI-system evaluation to drone propeller health monitoring. A higher policy-level AAS does not mean that a drone or propeller is chronologically older; rather, it indicates that a candidate scoring policy exhibits greater inconsistency or inadequacy under predefined metamorphic health-monitoring expectations.

The proposed framework separates three elements that are often conflated in simple scoring systems. First, the extracted flight-log indicators represent observed operational burden. Second, the selected scoring policy produces a propeller-health burden score. Third, the redundancy-adjusted AAS evaluates the structural adequacy of that scoring policy under metamorphic relations. This separation allows the prototype to report not only a maintenance recommendation, but also the dominant indicator, selected policy, policy-level AAS, dominant metamorphic violation, policy confidence, and decision confidence.

Metamorphic testing provides a suitable logic for this problem because an exact oracle for maintenance scoring is difficult to define. In software testing, the oracle problem arises when the expected output of a test case is unavailable or impractical to specify; metamorphic testing addresses this problem by checking whether relations between source and follow-up executions are preserved under defined transformations \cite{segura2016}. This relational logic has also been applied beyond conventional software testing in reliability-relevant settings where system behaviour must be evaluated through expected transformations rather than a single output value \cite{chen2018}. In the present study, metamorphic adequacy testing is adapted to drone propeller health monitoring by evaluating whether candidate scoring policies respond appropriately when trajectory tracking error, motor-command imbalance, ESC-command instability, or battery-level stress are systematically transformed.

The main contribution of this paper is the integration of DronePropA-based feature extraction, metamorphic adequacy testing, and redundancy-adjusted AAS modelling within a single interpretable decision-support prototype. The study defines a reproducible input structure from real flight-log data, evaluates candidate scoring policies under metamorphic relations, and adapts AAS as a structural policy-adequacy measure rather than as a chronological aging metric. In doing so, the proposed prototype extends AAS from behavioural or system-aging analysis to flight-log-based maintenance decision support for drone propeller health monitoring.
\section{Methodology}

\subsection{Research Design}

This study develops a retrospective decision-support prototype for drone propeller health monitoring using historical real flight-log data from the DronePropA public dataset. The aim is not to construct a conventional fault classifier, but to examine whether flight-log-derived indicators can be transformed into interpretable maintenance-support outputs through a Metamorphic Artificial Age Score (AAS) framework. The methodological structure combines four components: DronePropA-based feature extraction, healthy-baseline normalization, candidate scoring policies, and redundancy-adjusted AAS evaluation through metamorphic adequacy relations.

DronePropA is suitable for this purpose because it provides labelled flight-log data for healthy and defective drone propellers across multiple fault types, severity levels, speed profiles, and trajectory patterns \cite{ismail2024,ismail2025}. The dataset includes MATLAB \texttt{.mat} files containing \texttt{commander\_data}, \texttt{QDrone\_data}, and \texttt{stabilizer\_data}. In the present implementation, the six decision-support indicators are computed from the \texttt{commander\_data} and \texttt{QDrone\_data} matrices, while \texttt{stabilizer\_data} is part of the DronePropA file structure but is not used in the current feature set. This distinction is important because the reported indicators are derived from selected raw time-series signals rather than manually assigned or read as precomputed variables.

The evaluation is retrospective because the selected files are historical flight logs rather than live-streamed operational data. This design is appropriate for a proof-of-concept prototype because the purpose is to demonstrate whether real recorded flight behaviour can be transformed into structured maintenance recommendations. The resulting outputs are interpreted as post-flight decision-support evidence rather than as certified diagnostic or autonomous flight-control decisions.

\subsection{Selected DronePropA Subset}

A controlled subset of DronePropA was selected to keep the comparison internally consistent. All selected files correspond to the same speed profile and trajectory type. This reduces the risk that changes in the extracted indicators are caused by different trajectory demands or speed conditions rather than by propeller-health differences.

The selected files are:
\[
\begin{aligned}
&F0\_SV0\_SP1\_t1\_D1\_R1,\\
&F3\_SV1\_SP1\_t1,\\
&F3\_SV2\_SP1\_t1,\\
&F3\_SV3\_SP1\_t1.
\end{aligned}
\]

According to the DronePropA naming convention, \(F0\) denotes the healthy condition, while \(F1\), \(F2\), and \(F3\) denote fault groups. The severity codes \(SV1\), \(SV2\), and \(SV3\) represent increasing fault severity levels. The speed profile \(SP1\) corresponds to the higher-speed condition, and \(t1\) corresponds to the diagonal trajectory \cite{ismail2025}. Therefore, the selected subset consists of one healthy baseline and three defective propeller cases from the same fault group under the same speed and trajectory condition. This design allows the prototype to evaluate how extracted flight-log indicators change across a controlled healthy-to-defective sequence while avoiding unnecessary variation from speed or trajectory differences.
\subsection{DronePropA File Structure and Signal Extraction}

The data used in this study come from the public DronePropA dataset, originally released through Mendeley Data in 2024 and later described in a 2025 \textit{Data in Brief} article \cite{ismail2024,ismail2025}. DronePropA contains real flight-log and trajectory data for commercial drones operating with healthy and defective propellers. The dataset includes 130 flight sequences covering healthy and faulty propeller conditions, three fault types, three severity levels, two speed profiles, and five trajectory patterns. In the present study, a controlled subset of four MATLAB \texttt{.mat} files was selected from this dataset to compare one healthy baseline case with three defective cases under the same fault group, speed profile, and trajectory condition.

The numerical values used in this study were not manually assigned. They were extracted directly from the MATLAB \texttt{.mat} flight-log files using a Python implementation based on \texttt{scipy.io.loadmat}. For each selected file, the implementation loaded the \texttt{commander\_data} and \texttt{QDrone\_data} matrices. The \texttt{stabilizer\_data} matrix is part of the DronePropA file structure, but it was not used in the present feature set. MATLAB row numbers were converted into Python zero-based indexing during computation.

The signal-to-feature mapping used in the implementation is summarized in Table~\ref{tab:signal-mapping}. This table shows exactly which raw signals were extracted from the selected DronePropA files and how they were transformed into the six raw indicators used by the decision-support prototype.

\begin{table}[ht]
\centering
\caption{Signal-to-feature mapping used for DronePropA feature extraction.}
\label{tab:signal-mapping}
\resizebox{\textwidth}{!}{%
\begin{tabular}{llll}
\toprule
Feature & Source matrix & Rows used & Computation \\
\midrule
Trajectory tracking error & \texttt{commander\_data} & 22--24 and 26--28 & RMS of Euclidean error between measured and reference position \\
Attitude instability & \texttt{QDrone\_data} & 5--7 & Mean standard deviation and mean RMS temporal difference of roll, pitch, and yaw-rate signals \\
Thrust-command burden & \texttt{commander\_data} & 34 & RMS of temporal difference in reference thrust \\
Motor-command imbalance & \texttt{QDrone\_data} & 47, 49, 51, 53 & RMS of instantaneous standard deviation across four motor-command channels \\
ESC-command instability & \texttt{QDrone\_data} & 48, 50, 52, 54 & Average of ESC imbalance RMS and ESC temporal-difference RMS \\
Battery-level stress & \texttt{QDrone\_data} & 24 & Average of voltage drop and battery-level standard deviation \\
\bottomrule
\end{tabular}%
}
\end{table}

Accordingly, the raw feature table \(Z\) represents values computed directly from the selected flight-log signals. The normalized input table \(U\) is then obtained by comparing each raw feature value against the healthy baseline and retaining only positive deviations as burden evidence. Therefore, \(Z\) corresponds to the extracted raw flight-log indicators, while \(U\) corresponds to the baseline-normalized DSS input vectors used in the scoring, metamorphic adequacy, and AAS stages.

\subsection{Flight-Log-Derived Input Vector}

Each selected DronePropA file is transformed into a six-dimensional decision-support input vector:
\[
u=(u_1,u_2,u_3,u_4,u_5,u_6)\in[0,1]^6.
\]

The components are defined as follows:
\[
\begin{aligned}
u_1 &= \text{trajectory tracking error},\\
u_2 &= \text{attitude instability},\\
u_3 &= \text{thrust-command burden},\\
u_4 &= \text{motor-command imbalance},\\
u_5 &= \text{ESC-command instability},\\
u_6 &= \text{battery-level stress}.
\end{aligned}
\]

These variables are derived from the DronePropA \texttt{.mat} file structure. The \texttt{commander\_data} matrix provides measured and reference position signals, as well as reference thrust. The \texttt{QDrone\_data} matrix provides attitude-rate signals, battery level, motor-command signals, and electronic speed controller (ESC) command signals. The row indices used in the implementation follow the DronePropA variable description, with MATLAB one-based row numbers converted into Python zero-based indexing during computation.

The six indicators are intended to capture different operational channels through which propeller-health burden may appear. Trajectory tracking error represents deviation from the reference path. Attitude instability captures variation in roll, pitch, and yaw-rate behaviour. Thrust-command burden reflects variation in the reference thrust signal. Motor-command imbalance captures asymmetry across the four motor-command channels. ESC-command instability captures imbalance and temporal variation in ESC commands. Battery-level stress captures voltage drop and variation during the flight log.

\subsection{Raw Feature Extraction}

The trajectory tracking error is computed from the measured and reference position signals. Let
\[
p_m(t)=(x_m(t),y_m(t),z_m(t))
\]
represent the measured position and
\[
p_r(t)=(x_r(t),y_r(t),z_r(t))
\]
represent the reference position. The instantaneous position error is defined as
\[
e_p(t)=\|p_m(t)-p_r(t)\|_2.
\]
The raw trajectory tracking feature is computed as the root-mean-square value:
\[
z_1=\sqrt{\frac{1}{T}\sum_{t=1}^{T}e_p(t)^2}.
\]

Attitude instability is computed from roll-rate, pitch-rate, and yaw-rate signals. Let \(r_k(t)\) denote the \(k\)-th attitude-rate signal, where \(k\in\{1,2,3\}\). The feature combines the standard deviation and temporal derivative burden of these signals:
\[
z_2=
\frac{1}{2}\left(\frac{1}{3}\sum_{k=1}^{3}\sigma(r_k)\right)
+
\frac{1}{2}\left(\frac{1}{3}\sum_{k=1}^{3}RMS(\Delta r_k)\right).
\]

The thrust-command burden is computed from the reference thrust signal \(T_h(t)\):
\[
z_3=RMS(\Delta T_h).
\]

Motor-command imbalance is computed from the four motor-command channels. If \(m_j(t)\) denotes the command of motor \(j\), where \(j\in\{1,2,3,4\}\), the instantaneous motor-command imbalance is
\[
e_m(t)=\sigma(m_1(t),m_2(t),m_3(t),m_4(t)).
\]
The raw motor-command imbalance feature is then
\[
z_4=RMS(e_m(t)).
\]

ESC-command instability is computed from the four ESC command channels. If \(s_j(t)\) denotes the ESC command for channel \(j\), the instantaneous ESC-command imbalance is
\[
e_s(t)=\sigma(s_1(t),s_2(t),s_3(t),s_4(t)).
\]
The ESC-command instability feature combines ESC imbalance and temporal ESC variation:
\[
z_5=
\frac{1}{2}RMS(e_s(t))
+
\frac{1}{2}\left(\frac{1}{4}\sum_{j=1}^{4}RMS(\Delta s_j)\right).
\]

Battery-level stress is computed from the battery voltage signal \(b(t)\). The raw feature combines voltage drop and voltage variation:
\[
z_6=
\frac{1}{2}\max(0,b(1)-b(T))
+
\frac{1}{2}\sigma(b).
\]

The resulting raw feature vector is
\[
z=(z_1,z_2,z_3,z_4,z_5,z_6).
\]

\subsection{Baseline Normalization}

The raw feature values are normalized using the healthy file as the baseline. This produces a normalized vector in \([0,1]^6\), where \(0\) indicates no positive burden relative to the healthy baseline and \(1\) indicates the maximum observed positive burden within the selected retrospective subset.

Let \(z_{0,i}\) denote the healthy baseline value of feature \(i\). For each feature, the maximum positive deviation is defined as
\[
d_i^{\max}=\max_k(z_{k,i}-z_{0,i}).
\]
The normalized indicator is then computed as
\[
u_{k,i}=
\begin{cases}
0, & d_i^{\max}\leq 0,\\[6pt]
\min\left(1,\max\left(0,\dfrac{z_{k,i}-z_{0,i}}{d_i^{\max}}\right)\right), & d_i^{\max}>0.
\end{cases}
\]

This rule follows the burden interpretation used in the prototype: only positive deviations from the healthy baseline are treated as evidence of propeller-health burden. Decreases relative to the healthy condition are not interpreted as burden in this implementation. The normalization is retrospective and subset-based; therefore, broader operational deployment would require baseline modelling across multiple drones, speeds, trajectories, and operating environments.

\subsection{Candidate Scoring Policies}

The normalized input vector \(u\) is evaluated using three candidate scoring policies. The purpose of using multiple policies is not only to compute a propeller-health burden score, but also to evaluate which scoring policy behaves most adequately under metamorphic transformations.

The first policy is a linear weighted score:
\[
C_1(u)=\sum_{i=1}^{6}w_i u_i.
\]

The second policy is an operationally capped score:
\[
C_2(u)=\min\left(CAP,\sum_{i=1}^{6}w_i u_i\right).
\]
In the computational implementation, \(C_2\) also applies sequential rounding during accumulation to represent an operational scoring process in which values may be discretized.

The third policy is a threshold-sensitive score:
\[
C_3(u)=\sum_{i=1}^{6}w_i g_i(u_i),
\]
where \(g_i\) is an escalation function for selected indicators and the identity function for non-escalated indicators. In this prototype, trajectory tracking error, attitude instability, thrust-command burden, motor-command imbalance, and ESC-command instability are eligible for threshold-sensitive escalation, while battery-level stress is excluded. This design reflects the assumption that battery variation alone should not automatically dominate propeller-health interpretation.

The illustrative weight vector is
\[
w=(0.22,0.16,0.14,0.18,0.20,0.10),
\]
with
\[
\sum_{i=1}^{6}w_i=1.
\]
These weights are demonstration parameters. They are not universal constants and should be recalibrated through expert judgement, sensitivity analysis, and broader empirical validation before deployment.

\subsection{Metamorphic Adequacy Relations}

Metamorphic adequacy relations are used to evaluate the behaviour of each candidate scoring policy. The motivation is that a complete oracle for maintenance scoring is difficult to define. It may be unclear what exact score a flight log should receive, but it is possible to define relational expectations about how the score should behave under meaningful transformations. This follows the broader logic of metamorphic testing, where the relation between source and follow-up executions is used to evaluate correctness or adequacy when exact expected outputs are difficult to specify \cite{segura2016}.

Six metamorphic relations are used in this prototype:
\[
\begin{aligned}
MR_1 &= \text{uniform improvement},\\
MR_2 &= \text{tracking-error escalation},\\
MR_3 &= \text{attitude-instability escalation},\\
MR_4 &= \text{motor imbalance not masked},\\
MR_5 &= \text{ESC instability not masked},\\
MR_6 &= \text{battery alone non-critical}.
\end{aligned}
\]

The uniform improvement relation tests whether reducing all burden indicators fails to increase the score. The tracking-error escalation relation tests whether increasing trajectory tracking error produces an appropriate increase in score. The attitude-instability escalation relation tests whether increasing attitude instability produces an appropriate increase in score. The motor-imbalance relation tests whether increased motor-command imbalance is not fully masked by a decrease in battery stress. The ESC-instability relation tests whether increased ESC-command instability is not fully masked by a decrease in thrust-command burden. The battery-alone relation tests whether battery-level stress alone remains insufficient to trigger a critical propeller-health interpretation when the primary propeller-dynamics indicators are low.

For each policy \(C\) and metamorphic relation \(MR_r\), a source input \(u\) is transformed into a follow-up input \(u'_r\). The source and follow-up scores are
\[
y=C(u),
\qquad
y'_r=C(u'_r).
\]
A violation value \(v_r\geq0\) is then computed. A zero violation indicates that the policy satisfies the expected metamorphic relation. A positive violation indicates that the policy fails to satisfy the expected relation.

\subsection{Redundancy-Adjusted Artificial Age Score for Policy Adequacy}

The Artificial Age Score is used as a redundancy-adjusted policy adequacy measure. In the original AAS formulation, the score represents structural or behavioural burden through consistency loss, logarithmic penalty, and redundancy-aware aggregation rather than chronological time \cite{kayadibi2026}. In this study, the same logic is adapted to evaluate whether a drone propeller health scoring policy behaves adequately under metamorphic relations.

The AAS layer used here also builds on the redundancy-adjusted AAS formulation previously introduced for metamorphic testing \cite{kayadibi2026mt}. In that formulation, relation-level violation magnitudes are transformed into bounded consistency scores, mapped through a logarithmic penalty kernel, and aggregated with redundancy correction. The present study transfers this transformation from a general metamorphic-testing evaluation setting to a DronePropA-based drone propeller health-monitoring decision-support context. Therefore, AAS is not used merely to count violated metamorphic relations; rather, it converts relation-level inadequacy into a continuous, severity-sensitive, and redundancy-aware policy-adequacy measure.

Each metamorphic violation \(v_r\) is converted into a consistency score:
\[
x_r=\frac{1}{1+v_r}.
\]
Thus,
\[
x_r\in(0,1],
\]
where \(x_r=1\) indicates full consistency and lower values indicate greater violation. This step transforms the violation magnitude into a bounded consistency representation.

The logarithmic AAS penalty is then applied:
\[
\phi(x_r)=
-\log_2\left(\frac{x_r+\epsilon}{1+\epsilon}\right),
\qquad \epsilon>0.
\]
This penalty assigns zero burden when \(x_r=1\) and increases as consistency decreases. In this way, the transformation
\[
v_r \rightarrow x_r \rightarrow \phi(x_r)
\]
moves from a relation-level violation value to a penalty term that can be aggregated across metamorphic relations.

The redundancy-adjusted policy AAS is defined as
\[
AAS_C=
\sum_{r=1}^{m}
\omega_r(1-R_r)\phi(x_r),
\]
where \(m\) is the number of metamorphic relations, \(\omega_r\) is the relation weight, and \(R_r\in[0,1]\) is the redundancy adjustment for relation \(r\). In the present implementation, all metamorphic relations are assigned equal weights:
\[
\omega_r=\frac{1}{m}.
\]

The redundancy term \(R_r\) is operationalized from overlap among active violated metamorphic relations. Specifically, when a relation has no violation, its redundancy adjustment is set to zero. When it is violated, \(R_r\) is estimated from the average feature-set overlap between that relation and the other active violated relations. This implementation reflects the idea that overlapping violations should not necessarily be counted as fully independent evidence of policy inadequacy.

Accordingly, the full transformation
\[
v_r \rightarrow x_r \rightarrow \phi(x_r) \rightarrow AAS_C
\]
provides a higher-level aggregation of metamorphic adequacy evidence. The metamorphic relations identify whether a scoring policy violates expected health-monitoring behaviour under controlled input transformations, while the redundancy-adjusted AAS aggregates these relation-level violations into a policy-level adequacy score.

The selected policy is the one with the lowest redundancy-adjusted AAS:
\[
C^*=\arg\min_{C} AAS_C.
\]
Thus, the preferred policy is the one that produces the lowest redundancy-adjusted metamorphic inconsistency under the defined adequacy relations. However, this selection should be interpreted as the lowest-AAS policy within the tested candidate policy set, not as a globally optimal scoring policy.

\subsection{Maintenance Recommendation and Confidence Measures}

The selected scoring policy produces a propeller-health burden score:
\[
S=C^*(u).
\]

This score is used together with critical normalized indicators to generate a maintenance-support recommendation. The recommendation categories are:
\[
\text{Routine monitoring},
\quad
\text{Maintenance review / supervisory monitoring recommended},
\quad
\text{Mandatory inspection required}.
\]

The recommendation is based only on DSS-derived indicators. The retrospective severity label is not used to generate the recommendation. This ensures that the output depends on extracted flight-log evidence rather than known dataset labels.

Three indicators are treated as critical for escalation:
\[
u_1=\text{trajectory tracking error},
\qquad
u_4=\text{motor-command imbalance},
\qquad
u_5=\text{ESC-command instability}.
\]
Let
\[
K=\max(u_1,u_4,u_5)
\]
denote the critical indicator level. Mandatory inspection is triggered if the aggregate burden score exceeds the high threshold or if a critical indicator reaches a high normalized level. Maintenance review is recommended when the aggregate score or the critical indicator level indicates moderate burden. In the implementation, the low and high score thresholds are
\[
\theta_L=0.35,
\qquad
\theta_H=0.65.
\]
Mandatory inspection is also triggered when
\[
K\geq 0.90,
\]
and maintenance review is triggered when
\[
K\geq 0.50.
\]

The prototype also separates policy confidence from decision confidence. Policy confidence reflects the AAS margin between the best and second-best policies:
\[
M=AAS_{C_{\mathrm{second}}}-AAS_{C^*}.
\]
Decision confidence reflects the strength of the extracted indicators supporting the maintenance recommendation. This separation is important because candidate policies may be structurally close under the metamorphic AAS evaluation, producing weak policy confidence, while the maintenance recommendation may still be strongly supported by a dominant normalized indicator.

\subsection{Computational Implementation}

The computational implementation was written in Python. The workflow loads the selected DronePropA \texttt{.mat} files, extracts the required time-series signals from \texttt{commander\_data} and \texttt{QDrone\_data}, computes six raw flight-log-derived indicators, normalizes them relative to the healthy baseline, evaluates three candidate scoring policies, applies six metamorphic adequacy relations, computes redundancy-adjusted AAS values, and produces maintenance-support recommendations. The computational workflow is summarized below as Algorithm 1, Algorithm 2, and Algorithm 3.

The implementation produced five structured computational outputs: the raw feature table, normalized input table, policy-level AAS summary, policy-ranking details, and DSS decision summary. These outputs were used to construct the tables reported in the Results section. The implementation was used only as a retrospective decision-support prototype and not as certified diagnostic, maintenance-control, or autonomous flight-control software.

\subsubsection*{Algorithm 1. DronePropA feature extraction and baseline normalization}

\noindent\textbf{Input:} Selected DronePropA files
\[
\mathcal{F}=\{F_0,F_{SV1},F_{SV2},F_{SV3}\},
\]
where \(F_0\) is the healthy baseline file.\\
\textbf{Output:} Raw feature table \(Z\), where each row \(z_k=(z_{k,1},\ldots,z_{k,6})\) contains the six extracted flight-log indicators for case \(k\), and normalized DSS input table \(U\), where each row \(u_k=(u_{k,1},\ldots,u_{k,6})\in[0,1]^6\) contains the corresponding baseline-normalized burden indicators.

\begin{enumerate}[leftmargin=*, label=\arabic*.]
    \item Initialize an empty raw feature table \(Z\).

    \item For each selected file \(F_k\in\mathcal{F}\):
    \begin{enumerate}[leftmargin=*, label*=\arabic*.]
        \item Load the MATLAB \texttt{.mat} file.
        \item Read the \texttt{commander\_data} and \texttt{QDrone\_data} matrices.
        \item Extract measured position signals from rows 22--24 of \texttt{commander\_data}.
        \item Extract reference position signals from rows 26--28 of \texttt{commander\_data}.
        \item Compute the Euclidean position error:
        \[
        e_p(t)=\|p_m(t)-p_r(t)\|_2.
        \]
        \item Compute trajectory tracking error:
        \[
        z_{k,1}=RMS(e_p(t)).
        \]

        \item Extract roll-rate, pitch-rate, and yaw-rate signals from rows 5--7 of \texttt{QDrone\_data}.
        \item Compute attitude instability:
        \[
        z_{k,2}=
        \frac{1}{2}\left(\frac{1}{3}\sum_{i=1}^{3}\sigma(r_i)\right)
        +
        \frac{1}{2}\left(\frac{1}{3}\sum_{i=1}^{3}RMS(\Delta r_i)\right).
        \]

        \item Extract the reference thrust signal from row 34 of \texttt{commander\_data}.
        \item Compute thrust-command burden:
        \[
        z_{k,3}=RMS(\Delta T_h).
        \]

        \item Extract motor-command signals from rows 47, 49, 51, and 53 of \texttt{QDrone\_data}.
        \item Compute instantaneous motor-command imbalance:
        \[
        e_m(t)=\sigma(m_1(t),m_2(t),m_3(t),m_4(t)).
        \]
        \item Compute motor-command imbalance:
        \[
        z_{k,4}=RMS(e_m(t)).
        \]

        \item Extract ESC-command signals from rows 48, 50, 52, and 54 of \texttt{QDrone\_data}.
        \item Compute instantaneous ESC-command imbalance:
        \[
        e_s(t)=\sigma(s_1(t),s_2(t),s_3(t),s_4(t)).
        \]
        \item Compute ESC-command instability:
        \[
        z_{k,5}=
        \frac{1}{2}RMS(e_s(t))
        +
        \frac{1}{2}\left(\frac{1}{4}\sum_{j=1}^{4}RMS(\Delta s_j)\right).
        \]

        \item Extract the battery-level signal from row 24 of \texttt{QDrone\_data}.
        \item Compute battery-level stress:
        \[
        z_{k,6}
        =
        \frac{1}{2}\max(0,b(1)-b(T))
        +
        \frac{1}{2}\sigma(b).
        \]

        \item Store the raw feature vector:
        \[
        z_k=(z_{k,1},z_{k,2},z_{k,3},z_{k,4},z_{k,5},z_{k,6})
        \]
        in the raw feature table \(Z\).
    \end{enumerate}

    \item Sort the raw feature table \(Z\) by severity level.

    \item For each feature \(i\in\{1,\ldots,6\}\):
    \begin{enumerate}[leftmargin=*, label*=\arabic*.]
        \item Let \(z_{0,i}\) be the healthy baseline value for feature \(i\).
        \item Compute the deviation for each case:
        \[
        \delta_{k,i}=z_{k,i}-z_{0,i}.
        \]
        \item Compute the maximum positive deviation:
        \[
        d_i^{\max}=\max_k(\delta_{k,i}).
        \]
        \item If \(d_i^{\max}\leq TOL\), set
        \[
        u_{k,i}=0
        \]
        for all cases.
        \item Otherwise, compute the normalized DSS input:
        \[
        u_{k,i}
        =
        \min\left(1,\max\left(0,\frac{\delta_{k,i}}{d_i^{\max}}\right)\right).
        \]
    \end{enumerate}

    \item Return the raw feature table \(Z\) and normalized DSS input table \(U\).
\end{enumerate}
\subsubsection*{Algorithm 2. Metamorphic policy evaluation and redundancy-adjusted AAS calculation}

\noindent\textbf{Input:} Normalized DSS input table \(U\), candidate policies
\[
\mathcal{C}=\{C_1,C_2,C_3\},
\]
metamorphic relations
\[
\mathcal{M}=\{MR_1,\ldots,MR_6\},
\]
relation-feature sets, tolerance \(TOL\), and AAS parameter \(\epsilon>0\).\\
\textbf{Output:} Relation-level violation table and policy-level AAS ranking table.

\begin{enumerate}[leftmargin=*, label=\arabic*.]
    \item Define the candidate scoring policies.

    \item Define the linear policy:
    \[
    C_1(u)=\sum_{i=1}^{6}w_i u_i.
    \]

    \item Define the operationally capped policy:
    \[
    C_2(u)=
    \min\left(CAP,\sum_{i=1}^{6}w_i u_i\right),
    \]
    with sequential rounding during accumulation.

    \item Define the threshold-sensitive policy:
    \[
    C_3(u)=\sum_{i=1}^{6}w_i g_i(u_i),
    \]
    where selected indicators are transformed by
    \[
    g_i(v)=
    \begin{cases}
    0.85v, & v<\tau,\\
    \min(1,v+\beta(v-\tau)), & v\geq \tau.
    \end{cases}
    \]
    Battery-level stress is excluded from threshold escalation.

    \item Define the six metamorphic relations:
    \[
    MR_1=\text{uniform improvement},
    \]
    \[
    MR_2=\text{tracking-error escalation},
    \]
    \[
    MR_3=\text{attitude-instability escalation},
    \]
    \[
    MR_4=\text{motor imbalance not masked},
    \]
    \[
    MR_5=\text{ESC instability not masked},
    \]
    \[
    MR_6=\text{battery alone non-critical}.
    \]

    \item Construct the relation redundancy matrix \(Q\). For each pair of metamorphic relations \(MR_a\) and \(MR_b\), compute the Jaccard overlap between their feature sets:
    \[
    Q_{ab}=
    \frac{|F_a\cap F_b|}{|F_a\cup F_b|},
    \qquad a\neq b,
    \]
    and set \(Q_{aa}=0\).

    \item For each normalized case vector \(u_k\in U\):
    \begin{enumerate}[leftmargin=*, label*=\arabic*.]
        \item For each candidate policy \(C_j\in\mathcal{C}\):
        \begin{enumerate}[leftmargin=*, label*=\arabic*.]
            \item Compute the source score:
            \[
            y_j=C_j(u_k).
            \]

            \item For each metamorphic relation \(MR_r\in\mathcal{M}\):
            \begin{enumerate}[leftmargin=*, label*=\arabic*.]
                \item Generate the follow-up input:
                \[
                u'_{k,r}=MR_r(u_k).
                \]
                \item Compute the follow-up score:
                \[
                y'_{j,r}=C_j(u'_{k,r}).
                \]

                \item Compute the relation-specific violation value. For non-increase relations:
                \[
                v_{j,r}=\max(0,y'_{j,r}-y_j).
                \]
                For minimum-increase relations:
                \[
                v_{j,r}=\max(0,y_j+\gamma-y'_{j,r}).
                \]
                For the upper-bound relation:
                \[
                v_{j,r}=\max(0,y'_{j,r}-\theta_H).
                \]

                \item Convert the violation into a consistency score:
                \[
                x_{j,r}=\frac{1}{1+v_{j,r}}.
                \]

                \item Compute the logarithmic AAS penalty:
                \[
                \phi(x_{j,r})=
                -\log_2\left(\frac{x_{j,r}+\epsilon}{1+\epsilon}\right).
                \]

                \item Set the violation indicator:
                \[
                I_{j,r}=
                \begin{cases}
                0, & v_{j,r}\leq TOL,\\
                1, & v_{j,r}>TOL.
                \end{cases}
                \]

                \item Store the relation-level outputs:
                \[
                (y_j,y'_{j,r},v_{j,r},x_{j,r},\phi(x_{j,r}),I_{j,r}).
                \]
            \end{enumerate}

            \item For each relation \(MR_r\), compute its redundancy adjustment:
            \[
            R_{j,r}=
            \begin{cases}
            0, & I_{j,r}=0,\\
            0, & I_{j,r}=1 \text{ and no other relation is active},\\
            \mathrm{mean}(Q_{rs}: I_{j,s}=1,\ s\neq r), & I_{j,r}=1 \text{ and active overlaps exist}.
            \end{cases}
            \]

            \item Compute the redundancy-adjusted policy AAS:
            \[
            AAS_{C_j}
            =
            \frac{1}{m}
            \sum_{r=1}^{m}(1-R_{j,r})\phi(x_{j,r}).
            \]

            \item Store the policy-level outputs:
            \[
            \text{violation count},\quad
            \text{total violation},\quad
            \overline{R},\quad
            AAS_{C_j},\quad
            y_j.
            \]
        \end{enumerate}

        \item Rank the candidate policies by ascending policy AAS.
    \end{enumerate}

    \item Return the relation-level violation table and policy-level AAS ranking table.
\end{enumerate}
\subsubsection*{Algorithm 3. DSS recommendation, confidence, and dominant-output reporting}

\noindent\textbf{Input:} Normalized DSS input table \(U\), relation-level violation table, policy-level AAS ranking table, thresholds \(\theta_L=0.35\), \(\theta_H=0.65\), and critical indicator threshold \(0.90\).\\
\textbf{Output:} DSS decision summary.

\begin{enumerate}[leftmargin=*, label=\arabic*.]
    \item For each case \(k\):
    \begin{enumerate}[leftmargin=*, label*=\arabic*.]
        \item Select the best policy:
        \[
        C^*=\arg\min_{C_j\in\mathcal{C}} AAS_{C_j}.
        \]

        \item Let \(C_{\mathrm{second}}\) denote the second-ranked policy.

        \item Compute the AAS margin:
        \[
        M=AAS_{C_{\mathrm{second}}}-AAS_{C^*}.
        \]

        \item Assign policy confidence:
        \[
        \text{Policy confidence}=
        \begin{cases}
        \text{Strong}, & M\geq 0.025,\\
        \text{Moderate}, & 0.010\leq M<0.025,\\
        \text{Weak}, & M<0.010.
        \end{cases}
        \]

        \item Compute the selected propeller-health score:
        \[
        S_k=C^*(u_k).
        \]

        \item Compute the critical indicator level:
        \[
        K_k=\max(u_{k,1},u_{k,4},u_{k,5}),
        \]
        where \(u_{k,1}\) is trajectory tracking error, \(u_{k,4}\) is motor-command imbalance, and \(u_{k,5}\) is ESC-command instability.

        \item Assign the burden label:
        \[
        \text{Burden label}=
        \begin{cases}
        \text{Low}, & S_k<\theta_L,\\
        \text{Moderate}, & \theta_L\leq S_k<\theta_H,\\
        \text{High}, & S_k\geq \theta_H.
        \end{cases}
        \]

        \item Assign the maintenance recommendation:
        \[
        \text{Recommendation}=
        \begin{cases}
        \text{Mandatory inspection}, & S_k\geq \theta_H \text{ or } K_k\geq 0.90,\\
        \text{Maintenance review}, & S_k\geq \theta_L \text{ or } K_k\geq 0.50,\\
        \text{Routine monitoring}, & \text{otherwise}.
        \end{cases}
        \]

        \item Assign decision confidence. For mandatory inspection, decision confidence is strong if \(S_k\geq \theta_H\) or \(K_k\geq 0.90\). For maintenance review, decision confidence is strong if \(S_k\geq \theta_L\) or \(K_k\geq 0.65\). For routine monitoring, decision confidence is strong if \(K_k<0.30\).

        \item Determine the dominant indicator:
        \[
        \text{Dominant indicator}=
        \begin{cases}
        \text{None}, & \max_i u_{k,i}\leq TOL,\\
        \arg\max_i u_{k,i}, & \text{otherwise}.
        \end{cases}
        \]

        \item Determine the dominant metamorphic violation for the selected policy:
        \[
        \text{Dominant MR}=
        \begin{cases}
        \text{None}, & \max_r v_{C^*,r}\leq TOL,\\
        \arg\max_r v_{C^*,r}, & \text{otherwise}.
        \end{cases}
        \]

        \item Record the final DSS decision summary:
        \[
        \left(
        C^*,
        AAS_{C^*},
        M,
        \text{policy confidence},
        \text{decision confidence},
        S_k,
        \text{burden label},
        \text{recommendation},
        \text{dominant indicator},
        \text{dominant MR}
        \right).
        \]
    \end{enumerate}

    \item Return the DSS decision summary table.
\end{enumerate}
\section{Results}

\subsection{Raw Flight-Log-Derived Features}

The selected DronePropA files were processed to extract six raw flight-log-derived indicators: trajectory tracking error, attitude instability, thrust-command burden, motor-command imbalance, ESC-command instability, and battery-level stress. The extracted values are reported in Table~\ref{tab:raw-features}.

\begin{table}[ht]
\centering
\caption{Raw DronePropA-derived features.}
\label{tab:raw-features}
\resizebox{\textwidth}{!}{%
\begin{tabular}{lrrrrrr}
\toprule
Case & Tracking error & Attitude instability & Thrust-command burden & Motor-command imbalance & ESC-command instability & Battery stress \\
\midrule
Healthy & 0.504000 & 0.065206 & 0.126035 & 0.052673 & 0.012664 & 0.511987 \\
SV1     & 0.495916 & 0.064221 & 0.120589 & 0.053278 & 0.013578 & 0.495464 \\
SV2     & 0.491392 & 0.064220 & 0.118347 & 0.054866 & 0.014025 & 0.398089 \\
SV3     & 0.506982 & 0.063688 & 0.116697 & 0.053112 & 0.012734 & 0.506351 \\
\bottomrule
\end{tabular}%
}
\end{table}

The raw feature values show that the defective propeller cases do not produce a uniform increase across all flight-log channels. Trajectory tracking error decreases in SV1 and SV2 relative to the healthy baseline, but increases in SV3. Thrust-command burden also decreases across the defective cases. In contrast, motor-command imbalance and ESC-command instability increase in SV1 and SV2, with SV2 showing the highest values for both indicators. This pattern suggests that propeller fault effects are distributed across different operational channels rather than expressed through a single monotonically increasing raw signal.

\subsection{Normalized Decision-Support Inputs}

The raw features were normalized relative to the healthy baseline. Only positive deviations from the healthy case were treated as burden evidence. The resulting normalized decision-support inputs are reported in Table~\ref{tab:normalized-inputs}.

\begin{table}[ht]
\centering
\caption{Normalized DSS input vectors.}
\label{tab:normalized-inputs}
\resizebox{\textwidth}{!}{%
\begin{tabular}{lrrrrrr}
\toprule
Case & Tracking error & Attitude instability & Thrust-command burden & Motor-command imbalance & ESC-command instability & Battery stress \\
\midrule
Healthy & 0.000000 & 0.000000 & 0.000000 & 0.000000 & 0.000000 & 0.000000 \\
SV1     & 0.000000 & 0.000000 & 0.000000 & 0.275974 & 0.671929 & 0.000000 \\
SV2     & 0.000000 & 0.000000 & 0.000000 & 1.000000 & 1.000000 & 0.000000 \\
SV3     & 1.000000 & 0.000000 & 0.000000 & 0.200121 & 0.051229 & 0.000000 \\
\bottomrule
\end{tabular}%
}
\end{table}

The healthy case produced a zero burden vector:
\[
u_{\mathrm{Healthy}}=(0,0,0,0,0,0).
\]
The SV1 case produced non-zero burden mainly in ESC-command instability and motor-command imbalance:
\[
u_{\mathrm{SV1}}=(0,0,0,0.275974,0.671929,0).
\]
The SV2 case reached maximum normalized burden in both motor-command imbalance and ESC-command instability:
\[
u_{\mathrm{SV2}}=(0,0,0,1.000000,1.000000,0).
\]
The SV3 case reached maximum normalized burden in trajectory tracking error:
\[
u_{\mathrm{SV3}}=(1.000000,0,0,0.200121,0.051229,0).
\]

These normalized results show that the three defective propeller cases are not distinguished by the same dominant indicator. SV1 is primarily associated with ESC-command instability, SV2 with motor-command and ESC-command burden, and SV3 with trajectory tracking error. This supports the need for a multi-channel decision-support structure rather than a single-feature diagnostic rule.

\subsection{Policy-Level AAS Results}

The normalized DSS inputs were evaluated using three candidate scoring policies, \(C_1\), \(C_2\), and \(C_3\). Each policy was tested using six metamorphic adequacy relations, and the redundancy-adjusted AAS was computed for each case. The policy-level AAS results are reported in Table~\ref{tab:policy-aas}.

\begin{table}[ht]
\centering
\caption{Policy-level redundancy-adjusted AAS results.}
\label{tab:policy-aas}
\begin{tabular}{lrrr}
\toprule
Case & \(C_1\) & \(C_2\) & \(C_3\) \\
\midrule
Healthy & 0.000000 & 0.000000 & 0.000000 \\
SV1     & 0.001313 & 0.001199 & 0.000000 \\
SV2     & 0.016544 & 0.016544 & 0.016544 \\
SV3     & 0.008272 & 0.008272 & 0.008272 \\
\bottomrule
\end{tabular}
\end{table}

The healthy baseline produced zero AAS for all candidate policies, indicating that no metamorphic adequacy violation was detected. For SV1, the threshold-sensitive policy \(C_3\) produced the lowest AAS:
\[
AAS_{C_3}=0.000000.
\]
For SV2 and SV3, all candidate policies produced equal AAS values. This indicates that the current policy set does not strongly separate candidate policies for these cases. Therefore, policy confidence remains weak, although the maintenance recommendation may still be strongly supported by the extracted normalized indicators.

\subsection{Policy Ranking and Metamorphic Violation Summary}

The detailed policy-ranking results are reported in Table~\ref{tab:policy-ranking}. These results show how each candidate policy performed in terms of policy AAS, source score, violation count, total violation, and mean redundancy.

\begin{table}[ht]
\centering
\caption{Policy ranking details.}
\label{tab:policy-ranking}
\resizebox{\textwidth}{!}{%
\begin{tabular}{llrrrrrr}
\toprule
Case & Policy & Rank & Policy AAS & Source score & Violation count & Total violation & Mean \(R\) \\
\midrule
Healthy & C1 & 1 & 0.000000 & 0.000000 & 0 & 0.000000 & 0.000000 \\
Healthy & C2 & 2 & 0.000000 & 0.000000 & 0 & 0.000000 & 0.000000 \\
Healthy & C3 & 3 & 0.000000 & 0.000000 & 0 & 0.000000 & 0.000000 \\
SV1 & C3 & 1 & 0.000000 & 0.181645 & 0 & 0.000000 & 0.000000 \\
SV1 & C2 & 2 & 0.001199 & 0.184000 & 1 & 0.005000 & 0.000000 \\
SV1 & C1 & 3 & 0.001313 & 0.184061 & 1 & 0.005474 & 0.000000 \\
SV2 & C1 & 1 & 0.016544 & 0.380000 & 2 & 0.070000 & 0.000000 \\
SV2 & C2 & 2 & 0.016544 & 0.380000 & 2 & 0.070000 & 0.000000 \\
SV2 & C3 & 3 & 0.016544 & 0.380000 & 2 & 0.070000 & 0.000000 \\
SV3 & C1 & 1 & 0.008272 & 0.266267 & 1 & 0.035000 & 0.000000 \\
SV3 & C2 & 2 & 0.008272 & 0.266000 & 1 & 0.035000 & 0.000000 \\
SV3 & C3 & 3 & 0.008272 & 0.259327 & 1 & 0.035000 & 0.000000 \\
\bottomrule
\end{tabular}%
}
\end{table}

For SV1, \(C_3\) produced zero metamorphic violation, while \(C_1\) and \(C_2\) each produced one violation. This indicates that the threshold-sensitive policy was structurally adequate for the SV1 burden pattern. For SV2, all policies produced two violations and the same total violation value. For SV3, all policies produced one violation and the same policy AAS. The mean redundancy value was zero across the selected cases because active violations did not overlap sufficiently under the defined relation-feature structure. These results show that the current candidate policies are structurally close for SV2 and SV3 under the defined metamorphic relation set.

\subsection{DSS Decision Summary}

The final DSS outputs are reported in Tables~\ref{tab:dss-decision-main} and~\ref{tab:dss-decision-dominant}. The recommendation was generated from the selected policy score and normalized indicators. The retrospective severity label was not used to determine the maintenance recommendation.

\begin{table}[ht]
\centering
\caption{DSS decision summary: selected policy, AAS, confidence, score, and recommendation.}
\label{tab:dss-decision-main}
\begin{tabular}{llrrllrl}
\toprule
Case & Policy & AAS & Margin & Policy conf. & Decision conf. & Score & Recommendation \\
\midrule
Healthy & C1 & 0.000000 & 0.000000 & Weak & Strong & 0.000000 & Routine monitoring \\
SV1     & C3 & 0.000000 & 0.001199 & Weak & Strong & 0.181645 & Maintenance review \\
SV2     & C1 & 0.016544 & 0.000000 & Weak & Strong & 0.380000 & Mandatory inspection \\
SV3     & C1 & 0.008272 & 0.000000 & Weak & Strong & 0.266267 & Mandatory inspection \\
\bottomrule
\end{tabular}
\end{table}

\begin{table}[ht]
\centering
\caption{DSS decision summary: burden level, dominant indicator, and dominant metamorphic violation.}
\label{tab:dss-decision-dominant}
\begin{tabular}{llllrr}
\toprule
Case & Burden label & Dominant indicator & Dominant MR & Dominant value & MR violation \\
\midrule
Healthy & Low & None & None & 0.000000 & 0.000000 \\
SV1     & Low & ESC-command instability & None & 0.671929 & 0.000000 \\
SV2     & Moderate & Motor-command imbalance & MR4 & 1.000000 & 0.035000 \\
SV3     & Low & Trajectory tracking error & MR2 & 1.000000 & 0.035000 \\
\bottomrule
\end{tabular}
\end{table}

The healthy baseline was assigned to routine monitoring with strong decision confidence, consistent with the zero normalized burden vector. SV1 was assigned to maintenance review because ESC-command instability reached \(u_5=0.671929\), even though the aggregate propeller-health score remained low. No dominant metamorphic violation was reported for Healthy or SV1 because the selected policies produced zero relation-level violation for these cases.

SV2 was assigned to mandatory inspection because motor-command imbalance and ESC-command instability both reached the maximum normalized value:
\[
u_4=1.000000,\qquad u_5=1.000000.
\]
The selected policy for SV2 was \(C_1\), with a policy AAS of \(0.016544\), a source score of \(0.380000\), and dominant metamorphic violation \(MR_4\), indicating that motor-command imbalance was the most relevant violated adequacy relation.

SV3 was also assigned to mandatory inspection because trajectory tracking error reached the maximum normalized value:
\[
u_1=1.000000.
\]
The selected policy for SV3 was \(C_1\), with a policy AAS of \(0.008272\), a source score of \(0.266267\), and dominant metamorphic violation \(MR_2\), indicating that tracking-error escalation was the most relevant violated adequacy relation. Thus, SV2 represents the strongest control-channel burden among the selected files, while SV3 represents the strongest trajectory-level burden.
\subsection{Policy and Decision Confidence}

The results separate policy confidence from decision confidence. Policy confidence was weak across the cases because the AAS margin between the best and second-best policies was small. In SV2 and SV3, the candidate policies produced equal AAS values, so the selected policy should not be interpreted as uniquely superior.

Decision confidence was strong because the maintenance recommendations were supported by clear normalized indicators. The healthy case had no positive burden indicators. SV1 had elevated ESC-command instability. SV2 had maximum motor-command imbalance and ESC-command instability. SV3 had maximum trajectory tracking error. Therefore, a weak policy-selection margin does not imply a weak maintenance recommendation. It indicates that the candidate policies are structurally similar under the current metamorphic relation set, while the decision itself remains supported by the extracted flight-log indicators.

\subsection{Summary of Findings}

The results demonstrate that the proposed prototype can transform selected DronePropA flight-log data into interpretable maintenance-support outputs. The healthy baseline was assigned to routine monitoring, while the defective propeller cases were assigned to maintenance review or mandatory inspection depending on their dominant normalized indicators.

The main empirical finding is that propeller fault effects were not expressed through a single monotonically increasing indicator. SV1 was dominated by ESC-command instability, SV2 by motor-command imbalance and ESC-command instability, and SV3 by trajectory tracking error. The AAS layer provided an additional policy-adequacy interpretation by identifying how candidate scoring policies behaved under metamorphic transformations. Although policy confidence remained weak because the candidate policies were structurally close, decision confidence remained strong because the recommendations were supported by clear dominant indicators.
\section{Discussion}

\subsection{Interpretation of the DronePropA-Based Evaluation}

The results show that the proposed Metamorphic Artificial Age Score (AAS) decision-support prototype can transform selected DronePropA flight logs into interpretable maintenance-support outputs. The healthy baseline produced a zero normalized burden vector and was assigned to routine monitoring. In contrast, the defective propeller cases produced non-zero burden patterns and were assigned either to maintenance review or mandatory inspection. This demonstrates that the prototype can move from raw flight-log signals to structured decision-support outputs through a transparent computational pathway.

A central finding is that propeller fault effects were not expressed through a single monotonically increasing indicator. The SV1 case was dominated by ESC-command instability, the SV2 case by motor-command imbalance and ESC-command instability, and the SV3 case by trajectory tracking error. Thus, increasing labelled severity did not produce a uniform increase across all extracted indicators. Instead, different cases appeared through different operational channels. This is consistent with the motivation of DronePropA, where defective propeller conditions are represented through flight-log behaviour rather than through a single direct failure signal \cite{ismail2024,ismail2025}.

This finding supports the need for a multi-indicator decision-support approach. A system based only on trajectory tracking error would not capture the strongest burden pattern in SV1 and SV2. A system based only on motor-command imbalance would not fully capture the SV3 case. Similarly, a final aggregate score alone could obscure which operational channel was responsible for the maintenance recommendation. The proposed prototype addresses this by reporting the dominant indicator, propeller-health score, selected policy, metamorphic violation pattern, and maintenance recommendation.

\subsection{Role of the Artificial Age Score in the Proposed Prototype}

The Artificial Age Score is used in this study as a structural adequacy measure rather than as a chronological aging metric. The proposed framework does not interpret drone age as elapsed operational time or physical age of the propeller. Instead, it adapts the AAS logic to represent the structural burden and consistency behaviour of candidate scoring policies. This follows the original AAS interpretation, where artificial age is associated with behavioural or structural degradation, consistency loss, redundancy-adjusted burden, and logarithmic penalty rather than calendar time \cite{kayadibi2026}.

In the present prototype, AAS evaluates whether a candidate scoring policy behaves adequately under metamorphic health-monitoring expectations. When the input vector is transformed to represent uniform improvement, the score should not increase. When trajectory tracking error, motor-command imbalance, or ESC-command instability increases, the scoring policy should respond appropriately and should not allow these indicators to be fully masked by less relevant changes. The redundancy-adjusted AAS converts violations of such expectations into a structural inconsistency score.

This use of AAS separates two layers of interpretation. The propeller-health score summarizes the burden represented by the extracted flight-log indicators, whereas the policy-level AAS evaluates whether the scoring mechanism behaves consistently under the defined metamorphic relations. Therefore, AAS does not replace the maintenance score; it evaluates the adequacy of the scoring mechanism that produces it. This distinction gives the framework a stronger interpretive structure than a simple weighted scoring model.

\subsection{Value of Metamorphic Adequacy Testing}

Metamorphic testing is valuable in this setting because an exact oracle for drone maintenance scoring is difficult to define. It is not always possible to state the exact numerical score that a given flight log should receive. However, it is possible to define relational expectations about how the score should change when the input is transformed in meaningful ways. This follows the logic of metamorphic testing, which addresses the oracle problem by checking relations between source and follow-up executions rather than relying only on exact expected outputs \cite{segura2016}.

In this study, metamorphic adequacy testing allows the prototype to evaluate whether the scoring policy behaves in a structurally meaningful way. A policy may produce a plausible final score while still behaving inadequately under transformations. For example, it may allow motor-command imbalance to be masked by a reduction in battery stress, or it may fail to respond sufficiently to an increase in trajectory tracking error. These behaviours matter because maintenance decision support should be robust to the structure of the input space, not only to the final score.

The results show that metamorphic adequacy testing provided additional interpretive information. In SV2, the dominant metamorphic violation was associated with the motor-imbalance-not-masked relation. In SV3, the dominant metamorphic violation was associated with the tracking-error-escalation relation. These relation-level outputs help explain not only what recommendation was made, but also which structural expectation was most relevant to the case.

\subsection{Policy Confidence and Decision Confidence}

The results also show why policy confidence and decision confidence should be separated. Policy confidence remained weak because the AAS margin between the best and second-best candidate policies was small. In SV2 and SV3, all candidate policies produced equal AAS values. This indicates that the present policy set did not strongly distinguish among \(C_1\), \(C_2\), and \(C_3\) for these cases. This is not necessarily a failure of the maintenance recommendation; rather, it shows that the candidate policies are structurally close under the current metamorphic relation set.

Decision confidence, however, was strong because the maintenance recommendations were supported by clear normalized indicators. The healthy case had no positive burden indicators. SV1 had elevated ESC-command instability. SV2 had maximum normalized motor-command imbalance and ESC-command instability. SV3 had maximum normalized trajectory tracking error. These dominant indicators provide direct evidence for the corresponding maintenance outputs.

This separation prevents overinterpretation of the selected policy. A weak policy-confidence value means that the selected policy is not clearly superior to the alternatives under the current AAS margin. It does not mean that the maintenance recommendation itself is weak. Conversely, strong decision confidence indicates that the recommendation is supported by the extracted flight-log indicators. This distinction is important because model-selection confidence and operational recommendation confidence are not the same concept.

\subsection{Practical and Methodological Implications}

The proposed prototype is practically relevant because it provides a structured pathway from flight-log analysis to maintenance prioritization. For drone operators, maintenance teams, and autonomous-system oversight workflows, the value of such a system is not only that it produces a score, but that it explains the reason for the recommendation. The prototype identifies whether the burden comes primarily from trajectory tracking error, motor-command imbalance, ESC-command instability, or another channel.

This is particularly useful for post-flight analysis. A drone fleet may generate many flight logs, and maintenance teams may need to decide which vehicles or propeller sets require closer inspection. A binary healthy/faulty label may not be sufficient for this purpose. A decision-support output that identifies the dominant burden indicator and recommendation level can support more transparent prioritization.

The proposed framework is complementary to conventional fault-detection approaches. Prior UAV health-monitoring studies have shown the value of vibration-based and artificial-intelligence-based methods for identifying propeller or rotor abnormalities \cite{baldini2023,ghazali2022}. These approaches are useful for detection and classification. The present study addresses a different layer of the problem: how extracted indicators can be organized into an interpretable maintenance-support output. A classifier may indicate that a fault exists, while the AAS-DSS can help determine which indicator drives the burden, how the scoring policy behaves under metamorphic expectations, and whether the case should be assigned to routine monitoring, maintenance review, or mandatory inspection.

The results also have methodological implications. First, baseline normalization is important when using flight-log data because raw indicators have different units and scales. Normalization relative to the healthy baseline expresses each indicator as a comparable burden value in \([0,1]\). Second, positive deviations should be interpreted carefully. In this prototype, only positive deviations from the healthy baseline are treated as burden evidence. This is suitable for a proof-of-concept design, but broader deployment would require a more detailed model of expected operational variation. Third, the weak policy-confidence values suggest that future work should examine additional policy forms, adaptive weighting methods, nonlinear escalation rules, and richer metamorphic relation sets. Finally, the current redundancy values were zero because active violations did not overlap substantially under the defined relation-feature structure. This does not undermine the redundancy-adjusted formulation; rather, it shows that redundancy becomes informative only when overlapping metamorphic violations are active.

\subsection{Summary of Discussion}

Overall, the results support the value of a Metamorphic Artificial Age Score decision-support prototype for drone propeller health monitoring. The prototype transformed selected DronePropA flight logs into normalized indicators, identified dominant burden channels, evaluated candidate scoring policies through metamorphic adequacy relations, and generated interpretable maintenance recommendations. The main empirical insight is that propeller fault effects may appear through different operational channels rather than through a single monotonically increasing signal. The main methodological insight is that AAS can be adapted from behavioural or system-aging analysis to policy adequacy evaluation in drone maintenance decision support.

\section{Limitations and Future Work}

The proposed framework should be interpreted as a retrospective decision-support prototype. The evaluation demonstrates that selected DronePropA flight logs can be transformed into normalized propeller-health indicators and maintenance-support outputs, but broader validation is required before the framework can be considered for operational use. The selected subset includes one healthy baseline and three defective propeller cases from the same fault group, speed profile, and trajectory. This controlled design is useful for proof-of-concept evaluation, but it does not represent the full diversity of DronePropA, which includes multiple fault types, severity levels, speeds, trajectories, repetitions, and drones \cite{ismail2024,ismail2025}.

A second limitation concerns the normalization strategy. In the present prototype, the healthy baseline file is used as the reference condition, and only positive deviations from this baseline are treated as burden evidence. This makes the normalized indicators interpretable within the selected retrospective subset. However, broader deployment would require a more robust baseline model involving multiple healthy flights, different drones, different trajectories, and different speed conditions. Such a model would help separate normal operational variation from fault-related burden.

A third limitation concerns the candidate scoring policies. Policy confidence remained weak because the AAS margin between candidate policies was small in several cases. This indicates that \(C_1\), \(C_2\), and \(C_3\) are structurally close under the defined metamorphic relations. Future work should examine a wider range of scoring policies, adaptive weighting methods, nonlinear escalation rules, and data-informed threshold calibration. The aim would be to determine whether alternative policy structures can produce stronger separation while preserving interpretability.

A fourth limitation concerns the metamorphic adequacy relations. The six relations used in this prototype represent meaningful expectations for drone propeller health monitoring, including uniform improvement, tracking-error escalation, motor-imbalance sensitivity, ESC-instability sensitivity, and battery-alone non-critical behaviour. These relations provide a transparent starting point, but they are not exhaustive. Future research should expand the relation set to cover additional drone-health scenarios, including trajectory-specific behaviour, speed-dependent effects, thrust asymmetry, repeated-flight degradation, and interaction effects between motor, ESC, and attitude signals.

A fifth limitation concerns redundancy adjustment. In the current results, the mean redundancy value was zero because the active violation patterns did not overlap substantially under the defined relation-feature structure. This does not undermine the redundancy-adjusted AAS formulation, but it shows that redundancy becomes informative only when overlapping metamorphic violations are active. Larger evaluations across more DronePropA files may produce richer violation patterns and allow the redundancy component to play a stronger role.

Future work should extend the evaluation across the full DronePropA dataset. This would include all fault groups, all severity levels, both speed profiles, all trajectory types, and available healthy repetitions. A broader evaluation would allow the prototype to be tested for stability, sensitivity, and generalizability. It would also make it possible to compare AAS-DSS outputs with conventional fault-detection or classification models.

Further work should also examine whether the proposed framework can support post-flight fleet maintenance workflows. In this setting, the DSS could be used to rank flight logs according to maintenance priority, identify dominant burden indicators, and flag cases requiring closer inspection. The framework could also be extended toward near-real-time monitoring if suitable onboard processing and validated operational thresholds are available. However, such an extension would require additional validation, safety review, and domain-specific calibration.

Overall, the limitations point toward a clear development path. The present study establishes a proof-of-concept decision-support structure using selected drone flight-log data. Future work should focus on larger-scale empirical validation, expert-informed calibration, richer metamorphic relation design, and comparison with existing UAV fault-monitoring approaches.

\section{Conclusion}

This paper proposed a Metamorphic Artificial Age Score (AAS) decision-support prototype for flight-log-based drone propeller health monitoring. The framework transforms selected DronePropA flight logs into six normalized health-related indicators: trajectory tracking error, attitude instability, thrust-command burden, motor-command imbalance, ESC-command instability, and battery-level stress. These indicators are evaluated using candidate scoring policies, metamorphic adequacy relations, and a redundancy-adjusted AAS formulation.

The study used a controlled retrospective subset of DronePropA consisting of one healthy baseline and three defective propeller cases from the same fault group, speed profile, and trajectory. The healthy baseline produced a zero normalized burden vector and was assigned to routine monitoring. The Severity 1 case was dominated by ESC-command instability and was assigned to maintenance review. The Severity 2 case reached maximum normalized motor-command imbalance and ESC-command instability, triggering mandatory inspection. The Severity 3 case reached maximum normalized trajectory tracking error and also triggered mandatory inspection.

The results show that propeller fault effects may appear through different operational channels rather than through a single monotonically increasing indicator. This finding supports the use of a multi-indicator decision-support structure for drone propeller health monitoring. The proposed framework provides more than an aggregate burden score: it reports the dominant indicator, selected scoring policy, policy-level AAS, dominant metamorphic violation, policy confidence, and decision confidence.

AAS is used in this study as a structural adequacy measure rather than as a chronological aging metric. In this context, AAS evaluates the consistency of candidate scoring policies under metamorphic health-monitoring expectations. This allows the framework to connect flight-log-derived indicators with policy adequacy and maintenance prioritization.

The main contribution of the study is the integration of DronePropA-based feature extraction, metamorphic adequacy testing, and redundancy-adjusted AAS modelling within a single interpretable decision-support prototype. With broader validation and calibration, this approach may support transparent post-flight maintenance prioritization and autonomous-system oversight workflows for drone propeller health monitoring.

\section*{Data Availability Statement}

This study uses the publicly available DronePropA dataset. The original dataset is available through Mendeley Data as \textit{DronePropA: Motion Trajectories Dataset for Commercial Drones with Defective Propellers} \cite{ismail2024}. The accompanying data article is published in \textit{Data in Brief} \cite{ismail2025}. The present study used a controlled subset of selected MATLAB \texttt{.mat} flight-log files from this dataset.

\section*{Funding}

This research received no external funding.

\section*{Ethics Statement}

This study uses a publicly available drone flight-log dataset and does not involve human participants, animals, social media data, or personally identifiable information.

\section*{Code Availability Statement}

The Python implementation used for feature extraction, baseline normalization, metamorphic adequacy testing, redundancy-adjusted AAS calculation, and DSS output generation is available from the author upon reasonable request. The code was used to generate the structured computational outputs reported in the Results section, including the raw feature table, normalized input table, policy-level AAS summary, policy-ranking details, and DSS decision summary. The implementation is provided as a retrospective research prototype and should not be interpreted as certified diagnostic, maintenance-control, or autonomous flight-control software.


\begin{thebibliography}{99}

\bibitem{ismail2025}
Ismail, M. A. A., Elshaar, M. E., Abdallah, A., \& Quan, Q.
DronePropA: Motion trajectories dataset for defective drones.
\textit{Data in Brief}, \textit{60}, 111589, 2025.
\url{https://doi.org/10.1016/j.dib.2025.111589}

\bibitem{ismail2024}
Ismail, M. A. A., Elshaar, M. E., Abdallah, A., \& Quan, Q.
DronePropA: Motion Trajectories Dataset for Commercial Drones
with Defective Propellers.
Mendeley Data, V1, 2024.
\url{https://doi.org/10.17632/ftdyxrr3c5.1}

\bibitem{baldini2023}
Baldini, A., Felicetti, R., Ferracuti, F., Freddi, A.,
Iarlori, S., \& Monteri\`u, A.
Real-time propeller fault detection for multirotor drones
based on vibration data analysis.
\textit{Engineering Applications of Artificial Intelligence},
\textit{123}, 106343, 2023.
\url{https://doi.org/10.1016/j.engappai.2023.106343}

\bibitem{ghazali2022}
Ghazali, M. H. M., \& Rahiman, W.
Vibration-based fault detection in drone using artificial intelligence.
\textit{IEEE Sensors Journal}, \textit{22}(9), 8439--8448, 2022.
\url{https://doi.org/10.1109/JSEN.2022.3163401}

\bibitem{segura2016}
Segura, S., Fraser, G., Sanchez, A. B., \& Ruiz-Cort\'es, A.
A survey on metamorphic testing.
\textit{IEEE Transactions on Software Engineering},
\textit{42}(9), 805--824, 2016.
\url{https://doi.org/10.1109/TSE.2016.2532875}

\bibitem{chen2018}
Chen, T. Y., Kuo, F.-C., Liu, H., Poon, P.-L., Towey, D.,
Tse, T. H., \& Zhou, Z. Q.
Metamorphic testing: A review of challenges and opportunities.
\textit{ACM Computing Surveys}, \textit{51}(1), Article 4, 2018.
\url{https://doi.org/10.1145/3143561}

\bibitem{kayadibi2026}
Kayadibi, S. Y.
Redundancy-as-Masking: Formalizing the Artificial Age Score (AAS)
to Model Memory Aging in Generative AI.
\textit{Frontiers in Artificial Intelligence}, \textit{9}, 1732691, 2026.
\url{https://doi.org/10.3389/frai.2026.1732691}

\bibitem{kayadibi2026mt}
Kayadibi, S. Y.
Beyond Violation Counts: A Hypothesis-Driven Redundancy-Adjusted
Artificial Age Score for Metamorphic Testing.
\textit{Research Square preprint}, 2026.
\url{https://doi.org/10.21203/rs.3.rs-9530403/v1}

\end{thebibliography}
\end{document}